\documentclass[conference]{IEEEtran}
\IEEEoverridecommandlockouts
\usepackage{cite}
\usepackage{amsmath,amssymb,amsfonts}
\usepackage{algorithmic}
\usepackage{graphicx}
\usepackage{textcomp}
\usepackage{xcolor}
\usepackage{hyperref}

\PassOptionsToPackage{dvipsnames,svgnames,x11names}{xcolor} 
\usepackage{tikz} 

\usepackage{subfigure} 
\usepackage[numbers,sort&compress]{natbib} 
\usepackage{multirow} 
\usepackage{float} 
\usepackage{colortbl} 
\usepackage{makecell} 
\usepackage{bbding}  
\usepackage[linesnumbered,boxed,ruled,commentsnumbered]{algorithm2e} 
\usepackage{booktabs} 
\usepackage{pifont}  

\definecolor{YellowOrange}{rgb}{1.0, 0.68, 0.26}

\def\BibTeX{{\rm B\kern-.05em{\sc i\kern-.025em b}\kern-.08em
    T\kern-.1667em\lower.7ex\hbox{E}\kern-.125emX}}

\begin{document}\sloppy

\title{Multi-Level Correlation Network For Few-Shot Image Classification
}

\author{
\bf Yunkai Dang\textsuperscript{1}\quad Min Zhang\textsuperscript{2}\quad Zhengyu Chen\textsuperscript{2}\quad Xinliang Zhang\textsuperscript{1}\quad \\ 
\bf Zheng Wang\textsuperscript{1}\quad Meijun Sun\textsuperscript{1}\quad Donglin Wang\textsuperscript{2,$\dagger$}%
\thanks{$\dagger$ Corresponding author. Email: wangdonglin@westlake.edu.cn.}%
\thanks{This paper was completed at Westlake University. I would like to express my sincere gratitude to Min Zhang for her valuable suggestions, additional experiments, and revisions to this paper.}
\\
\textsuperscript{1}College of Intelligence and Computing, Tianjin University \\
\textsuperscript{2}AI Division, School of Engineering, Westlake University \\
}

\maketitle

\begin{abstract}
Few-shot image classification(FSIC) aims to recognize novel classes given few labeled images from base classes. Recent works have achieved promising classification performance, especially for metric-learning methods, where a measure at only image feature level is usually used. In this paper, we argue that measure at such a level may not be effective enough to generalize from base to novel classes when using only a few images. Instead, a multi-level descriptor of an image is taken for consideration in this paper. We propose a multi-level correlation network (MLCN) for FSIC to tackle this problem by effectively  capturing local information. Concretely, we present the self-correlation module and cross-correlation module to learn the semantic correspondence relation of local information based on learned representations. Moreover, we propose a pattern-correlation module to capture the pattern of fine-grained images and find relevant structural patterns between base classes and novel classes. Extensive experiments and analysis show the effectiveness of our proposed method on four widely-used FSIC benchmarks.
The code for our approach is available at: \href{https://github.com/Yunkai696/MLCN}{https://github.com/Yunkai696/MLCN}.
\end{abstract}


\section{Introduction}
Inspired by the human ability to recognize novel concepts from few images, few-shot image classification (FSIC) has attracted the interest of many researchers. FSIC aims to classify unlabeled images from novel classes given only few labeled images based on the trained model from base classes~\cite{finn2017model,matchingnet,snell2017prototypical,RelationNet,koch2015siamese}. 
Compared with traditional image classification tasks, the biggest challenge of FSIC is that the label spaces of base and novel classes are inconsistent, \textit{i.e.}, the labels of novel classes are not seen by base classes~\cite{li2021libfewshot}.
Most works have been proposed to solve the problem and are categorized, mainly divided into fine-tuning, meta-learning and metric-learning methods.
\textbf{Fine-tuning methods}~\cite{closer,tadam} use non-episode training paradigm (pre-training + test-tuning) to learn novel classes.
\textbf{Meta-learning methods}~\cite{chen2021meta,finn2017model} use the bi-level optimization with episodic-training paradigm.
\textbf{Metric-learning methods}~\cite{matchingnet, snell2017prototypical,RelationNet} also use episodic training but use distances to recognize novel classes in the embedding space.
In this paper, we consider metric-learning method because it is simple and effective.

\begin{figure}[t]
\centering
\includegraphics[width=0.9\linewidth]{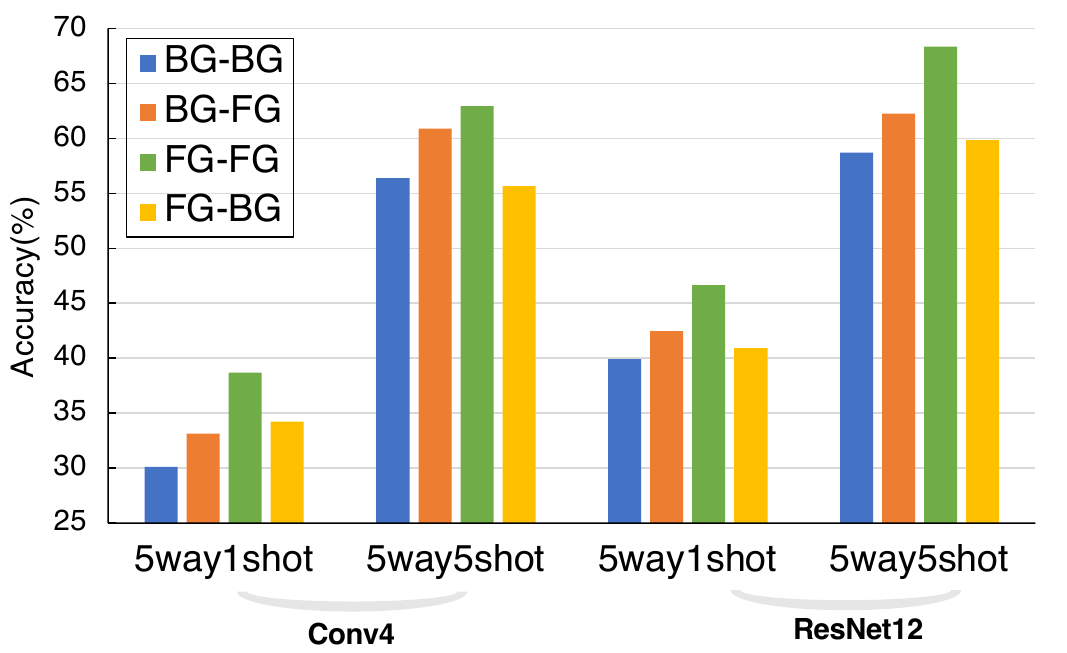}
\vspace{-10pt}
\caption{5way 1shot and 5shot performance on different background(BG) and foreground(FG) using two backbones of Conv4 and ResNet12 ProNet on \textit{mini}Imagenet datasets.}
\label{fig:motivation}
\end{figure}

Recently, metric-learning methods have achieved excellent performance by comparing the similarities (or distances) between the unlabeled (query set) and labeled (support set) images in the embedding space.
However, the similarity is calculated using only the global information of images. 
If base and novel classes have the distribution shift, the performance of the trained model degrades.
The main reason is that images from the same class tend to share similar backgrounds, which may cause them to be close to each other in the metric space. 
The transferability of the model trained on base classes becomes worse, when novel classes have different backgrounds from the base classes. 
We give a example in Fig.\textcolor{red}{~\ref{fig:motivation}} to show our motivation that the background is harmful.
It can be observed that under any condition, removing the background consistently leads to improved performance.
To alleviate the influence of image background, the main challenge of metric-learning methods is how to effectively capture the local information, such as the foreground of the image, on top of the global information.

To solve the challenge, in this paper, we propose a multi-correlation network (MLCN) to learn enough local information. Specifically, we present the self-correlation module and cross-correlation module to learn the semantic correspondence relations of local information based on the learned representation. Inspired by the way that human beings tend to instinctively focus on the most relevant areas of a pair of labeled and unlabeled images when trying to recognize a sample from an unseen class using only a few images~\cite{CAN}, 
we proposed a pattern-correlation module to capture the regions of fine-grained images and find relevant structural patterns between base classes and novel classes. The proposed MLCN combines the three correlation modules to improve the transferability of learned representations and learn enough local information to generalize to novel classes. Experiments on four standard benchmark datasets demonstrate that the proposed MLCN can effectively improve FSIC accuracy. 

\textbf{Contributions.} To summarize, our contributions are:
\begin{itemize}
    \item We verify that removing background consistently leads to the performance consistently and significantly improved. By taking this into consideration, we propose a multi-level correlation network (MLCN).
    \item We present the self-correlation module and cross-correlation module to learn the semantic correspondence relation of local information. Moreover, we propose the pattern-correlation module to capture the relevant structure of fine-grained images. 
    \item Experiments and analysis on four standard benchmarks show that our method achieves the state of the art and ablation studies validate the effectiveness of three correlation modules.
\end{itemize}

\section{Related work}

\textbf{Few-shot image classification.}
Few-shot image classification 
(FSIC) aims to recognize unlabeled images from the novel classes given few labeled images by transferring the knowledge from the base classes.
To solve this problem, researchers have proposed various methods, such as fine-tuning, meta-learning, and metric-learning methods.
\textbf{Fine-tuning methods}~\cite{closer,tadam,shotfree} are also called non-episode methods. These methods generally follow the standard transfer learning procedure, consisting of two phases, \textit{i.e.}, pre-training with base classes and test-tuning with novel classes.
\textbf{Meta-learning methods}~\cite{chen2021meta,finn2017model} adopt a \textit{learning-to-learn} paradigm to transfer the knowledge from the base classes to the novel classes. 
\textbf{Metric-learning methods}~\cite{matchingnet, snell2017prototypical,RelationNet} employ a \textit{learning-to-compare} paradigm to learn representations that can be transferred between the base and novel classes.
In this paper, our proposed method belongs to the metric-learning methods.

\textbf{Metric-learning methods.}
Metric learning~\cite{matchingnet, snell2017prototypical,RelationNet} is the method of learning a distance metric for the input space of base classes from a collection of pairs of similar and dissimilar points. Prototypical Network~\cite{snell2017prototypical,chen2021meta} is widely used in metric-based methods for FSIC. It takes the center point of a support class as its prototype and conducts classification by comparing similarities between query instances and support prototypes. However, such a process does not take into account the similarities between the query and support embedding. The main idea of our work is to improve the transferability of embedding by computing the similarities of the query and support embedding.

\textbf{Multi-level correlation network.}
Recent works~\cite{kwon2021learning,kang2021renet} adopt self-similarity and cross-similarity as an intermediate feature transformation for a deep neural network and show that it helps the network learn an effective representation. CAN~\cite{CAN} proposes a cross attention module to find the semantic relevance between the query and support set. Unlike the previous works~\cite{woo2018cbam}, our multi-level correlation network (MLCN) directly uses the correlation tensor to refine the representation of the query and support set. Different from ~\cite{yang2020prototype} using the Expectation-Maximization algorithm to learn features, we use the Bi-level optimization with episodic-training paradigm. Different from~\cite{kang2021renet,CAN}, we do not use the projection module and the 4D convolution without additional parameters to avoid overfitting for FSIC. 

\begin{figure*}[t]
\centering
\includegraphics[width=0.9\linewidth]{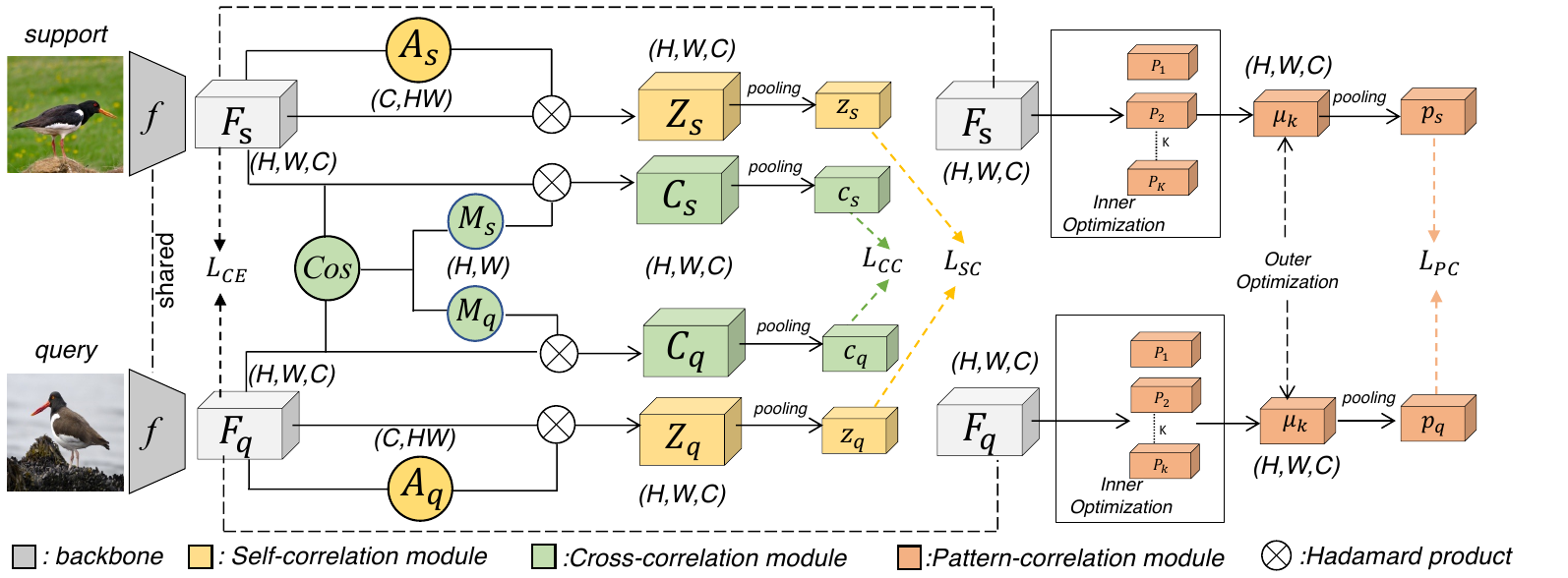}
\caption{
The overview of our multi-level correlation network (MLCN) for few-shot image classification. The base representations, $\mathbf{F}_{q}$ and $\mathbf{F}_{s}$ are the base representations of the backbone feature extractor. MLCN minimizes the self-correlation module loss $\mathcal{L}_{SC}$, the cross-correlation module loss $\mathcal{L}_{CC}$, the pattern-correlation module loss $\mathcal{L}_{PC}$ and the anchor-based classification loss $\mathcal{L}_{CE}$ to improve the transferability of embedding and capture enough local information.}
\label{fig:kuangjia}
\end{figure*}

\section{Methodology}
\subsection{Preliminary and Overview}
In the standard few-shot image classification (FSIC) scenario, a training data $\mathcal{D}_{train}$ from base classes $\mathcal{C}_{base}$ and a testing data $\mathcal{D}_{test}$ from novel classes $\mathcal{C}_{novel}$ are given, where $\mathcal{C}_{base} \bigcap \mathcal{C}_{novel} = \varnothing$. Following~\cite{CAN,kang2021renet}, we adopt the N-way K-shot episodic paradigm to train and test FSIC models, which has been demonstrated to be effective to enhance the generalization performance. For each episode, we randomly sample $N$ classes and $K$ labeled images per classes as the support set $\mathcal{S}=\left \{(x_{i}^{s}, y_{i}^{s}) |i=1,...,N \times K \right \} $ and a fraction of the remaning $Q$ images from the $N$ classes as the query set $\mathcal{Q}=\left \{ (x_{i}^{q}, y_{i}^{q}) |i=1,...,N \times Q \right \} $. FSIC aims to learn a classifier to classify images into target classes given only a few images for each class. The key issue is to present each support class and query sample and measure the similarity between them for few-shot image classification.

In this section, we propose the multi-level correlation network (MLCN) to address the challenge of generalization to novel classes from a similarity perspective.
As shown in Fig.\textcolor{red}{~\ref{fig:kuangjia}}, our MLCN has three complementary branches based on the backbone. Then we present technical details of self-correlation module, cross-correlation module and pattern-correlation module in Sec.\textcolor{red}{3.2}, Sec.\textcolor{red}{3.3} and Sec.\textcolor{red}{3.4}. Last, we introduce the overall loss in Sec.\textcolor{red}{3.5}.

\subsection{Self-correlation Module}
To improve the generalizability of the representation, as shown in Fig.\textcolor{red}{~\ref{fig:kuangjia}}, we propose the architecture of self-correlation module. Here, we use it to transform the base representations, $F_{q}$ and $F_{s}$ to self-correlation representations, $Z_{q}$ and $Z_{s}$ by the self-correlation attention map $A_{q}$ and $A_{s}$.

\textbf{Self-correlation Attention Map.}
To locate the discriminative object regions of query and support set, we firstly produce a self-correlation attention map, $A_{q} $ and $A_{s} \in R^{C \times HW}$, which generates the weights of base representations, $F_{q}$ and $F_{s} \in R^{H\times W\times C}$.
\begin{equation}\label{SC_Aq}
    \mathbf{A}_{q} \left ( x_{q} \right ) = softmax(F_{q}\left ( x_{q} \right )  ),    
\end{equation}
where $x_{q}$ is position at the query feature map and $A_{s}$ is similarly computed by  $\mathbf{E_{q.}}($\textcolor{red}{~\ref{SC_Aq}}).
The self-correlation embeddings, $\mathbf{Z}_{s}$ and $\mathbf{Z}_{q} \in R^{C}$, are multiplied by the self-correlation attention map $A_{q} $. Then, we compute the final self-correlation embedding $\mathbf{z}_{q}$ of query feature by using pooling:
\begin{equation}\label{SC_q}
      \mathbf{z_{q}}=\frac{1}{HW}\sum_{x_{q}}^{} A_{q}\left ( x_{q} \right )  \otimes  F_{q}  \left (  x_{q}\right ),
\end{equation}
where x denotes a position at the feature map. The final embedding $\mathbf{z_{s}}$ of support images is computed similarly by $\mathbf{E_{q.}}$(\textcolor{red}{~\ref{SC_q}}).
In the N-way K-shot classification setting, this self-correlation pooling generates a set of NK different views of a query, $\left \{\mathbf{\bar{z}_{q}}^{(n)} \right \}_{n=1}^{N}$, and a set of support embedding in the context of the query, $\left \{\mathbf{\bar{z}_{s}}^{(n)} \right \}_{n=1}^{N}$. 

\textbf{$\mathcal{L}_{SC}$ : self-correlation module loss.}
We average the K query and support embedding vectors, each of which is attended in the context of $k^{th}$ support from $n^{th}$ class to compute $\left \{\mathbf{\bar{z}_{q}}^{(n)} \right \}_{n=1}^{N}$ and $\left \{\mathbf{\bar{z}_{s}}^{(n)} \right \}_{n=1}^{N}$. The self-correlation module loss guides the model to map a query embedding close to the prototype embedding of the same class in the metric space, so we have
\begin{equation}\label{Loss_SC}
      \mathcal{L}_{\text {SC}}=-\log \frac{\exp
    \left(\operatorname{sim}\left(\mathbf{\bar{z}_{s}}^{(n)}, 
    \mathbf{\bar{z}_{q}}^{(n)}\right) / \tau_{1}\right)}{\sum_{n^{\prime}=1}^{N} 
    \exp \left(\operatorname{sim}\left(\mathbf{\bar{z}_{s}}^{\left(n^{\prime}\right)}, 
    \mathbf{\bar{z}_{q}}^{\left(n^{\prime}\right)}\right) / \tau_{1}\right)},
\end{equation}
where $\tau_{1}$ is a scalar temperature factor and $sim\left ( .,. \right )$ is a cosine similarity .

\subsection{Cross-correlation Module}
The self-correlation module localizes the discriminative object regions of each query and support set may ignore the semantic correspondence relation between them. To learn this relation, as shown in Fig.\textcolor{red}{~\ref{fig:kuangjia}}, we use the cross-correlation module to further find the reveal relevant contents between the query and support set.

\textbf{Cross-correlation Attention Map.}
To improve the transferability of embedding, we transform base representations $F_{s}$ and $F_{q}$ into more compact representations by constructing a 4-dimensional correlation tensor $\mathbf{Cos} \in R^{H\times W\times H \times W}$, which computes the cosine similarity between two features.
The cross-correlation attention map $A_{q} \in R^{H\times W}$ of the query is computed by
{\small \begin{equation}\label{CC_Aq}
       \mathbf{M}_{q}=\frac{1}{HW}\sum_{x_{s}}^{} \frac{\exp\left (Cos(F_{q}\left ( x_{q} \right ) ,F_{s}
     \left ( x_{s} \right )))/\gamma   \right ) }{\sum_{x_{q'}}^{} \exp\left (Cos(F_{q}\left ( x_{q{}' } \right ),F_{s}\left ( x_{s} \right ))
    /\gamma \right )  },   
\end{equation}{\small {\small }}}where $\gamma$ is a scalar temperature parameter and $Cos(.,.)$ is a matching score between the positions $x_{q}$ and $x_{s}$ of the $F_{q}$ and $F_{s}$. And the ${M}_{s}$ is similarly computed by $\mathbf{E_{q.}}$\textcolor{red}{(~\ref{CC_Aq})}. Similar to $\mathbf{E_{q.}}$\textcolor{red}{(~\ref{SC_q})}, we compute the cross-correlation of support feature embedding $c_{q}\in R^{H\times W\times C}$ by using the pooling:
\begin{equation}\label{CC_q}
      \mathbf{c_{q}}=\frac{1}{HW}\sum_{x_{q}}^{} M_{q}\left ( x_{q} \right )  \otimes  F_{q}  \left (  x_{q}\right ), 
\end{equation}
where $x_{q}$ is a position at the query feature map. The final embedding of the query, $\mathbf{c_{s}}$, is computed similarly by $\mathbf{E_{q.}}$ (\textcolor{red}{~\ref{CC_q}}).

\textbf{$\mathcal{L}_{CC}$ : cross-correlation module loss.}
Similar to $\mathbf{E_{q.}}$ (\textcolor{red}{~\ref{SC_q}}), we compute the support feature embedding by multiplying base representations $F_{s}$.
Similar to self-correlation module loss, we average the K query and support embedding vectors as $\left \{\mathbf{\bar{c}_{q}}^{(n)} \right \}_{n=1}^{N}$ and $\left \{\mathbf{\bar{c}_{s}}^{(n)} \right \}_{n=1}^{N}$. 
The cross-correlation module loss guides the model to map a query embedding close to the prototype embedding of the same class:

\begin{equation}\label{Loss_CC}
   \mathcal{L}_{\text {CC }}=-\log \frac{\exp
    \left(\operatorname{sim}\left(\mathbf{\bar{c}_{s}}^{(n)}, 
    \mathbf{\bar{c}_{q}}^{(n)}\right) / \tau_{2}\right)}{\sum_{n^{\prime}=1}^{N} 
    \exp \left(\operatorname{sim}\left(\mathbf{\bar{c}_{s}}^{\left(n^{\prime}\right)}, 
    \mathbf{\bar{c}_{q}}^{\left(n^{\prime}\right)}\right) / \tau_{2}\right)},
\end{equation}
where $sim(.,.)$ indicates the cosine similarity and $\tau_{2}$ is a scalar temperature factor. 

\subsection{Pattern-correlation Module }
As shown in Fig.\textcolor{red}{~\ref{fig:kuangjia}}, we use the pattern-correlation module to find relevant structural pattern between base classes and novel classes.
Pattern-correlation module is defined as a probability mixture model which linearly combines probabilities from base distributions as:
\begin{equation}\label{MP_p}
    p\left ( s_{i}|\theta   \right )=
 {\textstyle \sum_{k=1}^{K}} w_{k} p_{k}\left ( s_{i}|\theta   \right ),
\end{equation}
where $w_{k}$ denotes the mixing weights satisfying $0\le w_{k} \le 1$ and $ {\textstyle \sum_{k=1}^{K}} w_{k} =1$. $\theta$ denotes the model parameters which are learned by the backbone for extracting embedding. $s_{i} \in S$ denotes the $i^{th}$ feature samples and $p_{k}\left ( s_{i}|\theta   \right )$ denotes the $k^{th}$ base model. Thus, we calculate the $p_{k}\left ( s_{i}|\theta   \right )$ as :
\begin{equation}\label{MP_distance}
   p_{k}\left ( s_{i}|\theta   \right )=
\beta \left ( \theta  \right )
e^{distance},
\end{equation}
where $\beta \left ( \theta  \right )$ is the normalization coefficient, and the $distance$ is defined as the euclidean distance of $\mu _{k}$ and $s_{i}$.
Here, $\mu _{k}$ is the mean vector of the $k^{th}$ model, and $k$ denotes the concentration parameter.
Pattern-correlation module is estimated by using the bi-level optimization, which includes iterative inner optimization and outer optimization. In \textbf{inner optimization step}, we use the $ p_{k}\left ( s_{i}|\theta   \right )$ to extract sample features as 
\begin{equation}\label{MP_uk}
   \\P _{ik} = \frac{p_{k}\left ( s_{i}|\theta   \right ) }
{ {\textstyle \sum_{k=1}^{K}}p_{k}\left (  s_{i}|\theta  \right )  }, 
\end{equation}
where k denotes the concentration parameter and is set as k =25 in experiments. In \textbf{outer optimization}, we use the $P_{ik}$ to update the mean vectors as $u_{k} \in R^{H\times W\times C}$, which is computed by
\begin{equation}\label{MP_u}
   \mu _{k} = \frac{ {\textstyle \sum_{i=1}^{N}P_{ik}s_{i}}  }
{ {\textstyle \sum_{i=1}^{N}P_{ik}}  }, 
\end{equation}
where $s_{i}$ denotes the $i^{th}$ feature samples and  $N$ denotes the number of samples.

\textbf{$\mathcal{L}_{PC}$ : pattern-correlation module loss.}
The final embeddings of and query $\mathbf{p_{q}}$ and support $\mathbf{p_{s}}$ are computed by pooling of the $\mu _{k}$.
Similar to $\mathbf{E_{q.}}$ (\textcolor{red}{~\ref{Loss_CC}}) and  $\mathbf{E_{q.}}$ (\textcolor{red}{~\ref{Loss_SC}}), we average the K query and support embedding vectors for each class to obtain a set of prototype embedding $\left \{\mathbf{\bar{p_{q}}}^{n} \right \}_{n=1}^{N}$ and $\left \{\mathbf{\bar{p}_{s}}^{n} \right \}_{n=1}^{N}$. {$\mathcal{L}_{PC}$ is computed by cosine similarity between a query and support prototypes:
\begin{equation}\label{Loss_PC}
   \mathcal{L}_{\text {PC }}=-\log \frac{\exp
    \left(\operatorname{sim}\left(\mathbf{\bar{p}_{s}}^{(n)},
    \mathbf{\bar{p}_{q}}^{(n)}\right) / \tau_{3}\right)}{\sum_{n^{\prime}=1}^{N} 
    \exp \left(\operatorname{sim}\left({\mathbf{\bar{p}_{s}}}^{\left(n^{\prime}\right)}, 
    {\mathbf{\bar{p}_{q}}}^{\left(n^{\prime}\right)}\right) / \tau_{3}\right)},
\end{equation}
where $sim(.,.)$ is cosine similarity and $\tau_{3}$ is a scalar temperature factor. 
\subsection{Overall Loss }
the anchor-based classification loss $\mathcal{L}_{CE}$ is computed with an additional fully-connected classification layer on top of average-pooled based feature $F_{q}$. It guides the model to classify a query class of class $c\in C_{base}$, so we have
\begin{equation}\label{Loss_CE}
   \mathcal{L}_{\text {CE }}=-\log \frac{\exp \left(\mathbf{w}_{c}^{\top} \mathbf{F}_{\mathrm{q}}+\mathbf{b}_{c}\right)}{\sum_{c^{\prime}=1}^{\left|\mathcal{C}_{\text {base }}\right|} \exp \left(\mathbf{w}_{c^{\prime}}^{\top} \mathbf{F}_{\mathrm{q}}+\mathbf{b}_{c^{\prime}}\right)},
\end{equation}
where $\left [ \mathbf{w}_{1}^{T},..., \mathbf{w}_{|C_{base}|}^{T}\right ]$ and $\left [ \mathbf{b}_{1},..., \mathbf{b}_{|C_{base}|}\right ] $ are weights and biases in the fully-connected layer. 
In summary, the final loss of each episode is defined as :
\begin{equation}\label{Loss_all}
   \mathcal{L}_{\text {total}} = \mathcal{L}_{\text {CE}} 
+ \alpha \mathcal{L}_{\text {SC}} +\beta  \mathcal{L}_{\text {CC}}+ \gamma  \mathcal{L}_{\text {PC}}, 
\end{equation}
where, $\alpha$, $\beta$ and $\gamma$ are weighting factor to balance the loss terms.
\section{Experiment}

\begin{table*}[t]
\centering    
    \caption{Ablation experiments on \textit{mini}Imagenet, CIFAR-FS and CUB200-2011 datasets. $\mathcal{L}_{CE}$ is the anchor-based classification loss. $\mathcal{L}_{SC}$, $\mathcal{L}_{CC}$ and $\mathcal{L}_{PC}$ are the loss of the self-correlation module, cross-correlation module and patterns-correlation module.}
    \renewcommand\arraystretch{1.1}
    \resizebox{2\columnwidth}{!}{
\begin{tabular}{cccc|ll|ll|ll}
\hline
\multirow{2}{*}{$L_{CE}$} & \multirow{2}{*}{$L_{SC}$} & \multirow{2}{*}{$L_{CC}$} & \multirow{2}{*}{$L_{PC}$} & \multicolumn{2}{c|}{\textbf{miniImageNet}}& \multicolumn{2}{c|}{\textbf{CIFAR-FS}} & \multicolumn{2}{c}{\textbf{CUB200-2011}}\\ 
\cline{5-10} 
                          &                           &                       &                             & \multicolumn{1}{c|}{\textbf{1-shot}} & \multicolumn{1}{c|}{\textbf{5-shot}} & \multicolumn{1}{c|}{\textbf{1-shot}} & \multicolumn{1}{c|}{\textbf{5-shot}} & \multicolumn{1}{c|}{\textbf{1-shot}} & \multicolumn{1}{c}{\textbf{5-shot}} \\ \hline
$\surd$                   &                           &                       &   
& \multicolumn{1}{l|}{57.89 ± 0.44}    & 76.94 ± 0.34                         
& \multicolumn{1}{l|}{61.25 ± 0.48}    & 79.63 ± 0.34                         
& \multicolumn{1}{l|}{69.92 ± 0.43}    & 73.64 ± 0.36                        \\
$\surd$                   & $\surd$                   &                       &              
& \multicolumn{1}{l|}{64.31 ± 0.44}    & 81.11 ± 0.31                         
& \multicolumn{1}{l|}{71.19 ± 0.45}    & 86.37 ± 0.33
& \multicolumn{1}{l|}{ 73.91 ± 0.44}    & 87.33 ± 0.28  
\\
$\surd$                   & $\surd$                   & $\surd$               &                     & \multicolumn{1}{l|}{64.41 ± 0.45}    & 81.54 ± 0.31                      
& \multicolumn{1}{l|}{70.58 ± 0.46}    & 86.55 ± 0.32                         
& \multicolumn{1}{l|}{74.40 ± 0.67}    & 89.19 ± 0.27
\\ 
$\surd$                   & $\surd$                   &                       & $\surd$             & \multicolumn{1}{l|}{64.89 ± 0.44}    & 81.33 ± 0.30                         
& \multicolumn{1}{l|}{72.60 ± 0.46}    & 85.78 ± 0.33                         
& \multicolumn{1}{l|}{76.00 ± 0.45}    & 89.12 ± 0.26                        \\
$\surd$                   & $\surd$                   & $\surd$               & $\surd$             & \multicolumn{1}{l|}{\textbf{65.54 ± 0.44}}    & \textbf{81.94 ± 0.31}                         
& \multicolumn{1}{l|}{\textbf{74.36 ± 0.47}}    & \textbf{87.24 ± 0.31}                        
& \multicolumn{1}{l|}{\textbf{77.96 ± 0.44}}    & \textbf{91.20 ± 0.24}                        
\\ \hline
\end{tabular}
}
\end{table*}
\begin{table}[t!]
\caption{Results on the \textit{mini}Imagenet dataset.}
\centering
\tabcolsep 4pt
\small

\scalebox{0.98}{
\begin{tabular}{lccc}
\hline
\textbf{Mthod} & \textbf{Backbone} & \textbf{5-way 1-shot} & \textbf{5-way 5-shot}\\  \hline
Versa~\cite{gordon2018versa}  & \emph{ResNet12}  & 55.71  & 70.05 \\
LEO~\cite{leo}  & \emph{ResNet12}  & 56.62  & 69.99 \\
BOIL~\cite{oh2020boil}  & \emph{ResNet12}  & 58.87  & 72.88 \\
R2D2~\cite{bertinetto2018meta}  & \emph{ResNet12}  & 59.52  & 74.61 \\
MTL~\cite{mtl}  & \emph{ResNet12}  & 62.67  & 79.16 \\
\hline
TPN~\cite{tpn}  & \emph{ResNet12}  & 59.46    & 75.65    \\
CC~\cite{closer} & \emph{ResNet12}  & 55.43 $\pm$ 0.81 & 77.18 $\pm$ 0.61 \\
TapNet~\cite{CAN}  & \emph{ResNet12}  & 61.65 $\pm$ 0.15 & 76.36 $\pm$ 0.10\\
MetaOptNet~\cite{metaoptnet}  & \emph{ResNet12}  & 62.64 $\pm$ 0.82 & 78.63 $\pm$ 0.46\\
MatchNet~\cite{matchingnet}  & \emph{ResNet12}  & 63.08 $\pm$ 0.80 & 75.99 $\pm$ 0.60\\
ProtoNet~\cite{protonet}  & \emph{ResNet12}  & 62.39 $\pm$ 0.21  & 80.53 $\pm$ 0.14 \\ %
CAN~\cite{CAN}  & \emph{ResNet12}  & 63.85 $\pm$ 0.48 & 79.44 $\pm$ 0.34\\
RFS-simple~\cite{rfs} & \emph{ResNet12}  & 62.02 $\pm$ 0.63 & 79.64 $\pm$ 0.44\\
\hline
 \textbf{MLCN(ours)}  &  \emph{ResNet12}  &  \textbf{65.54 $\pm$ 0.43} &  \textbf{81.63 $\pm$ 0.31} \\ 
\hline
\end{tabular}
}
\end{table}

\begin{table}[t!]
\caption{  Results on the \textit{tiered}ImageNet dataset.
 }
\centering
\tabcolsep 4pt
\small

\scalebox{0.98}{
\begin{tabular}{lccc}
\hline
\textbf{Mthod} & \textbf{Backbone} & \textbf{5-way 1-shot} & \textbf{5-way 5-shot}\\  \hline
Versa~\cite{gordon2018versa}  & \emph{ResNet12}  & 57.14  & 75.48 \\
LEO~\cite{leo}  & \emph{ResNet12}  & 64.75  & 81.42 \\
BOIL~\cite{oh2020boil}  & \emph{ResNet12}  & 64.66  & 80.38 \\
R2D2~\cite{bertinetto2018meta}  & \emph{ResNet12}  & 65.07  & 83.04 \\
MTL~\cite{mtl}  & \emph{ResNet12}  & 68.68  & 84.58 \\
\hline
TPN~\cite{tpn}  & \emph{ResNet12}  & 59.91 $\pm$ 0.94  & 73.30 $\pm$ 0.75 \\
CC~\cite{closer} & \emph{ResNet12}  & 61.49 $\pm$ 0.91  & 82.37 $\pm$ 0.67 \\
TapNet~\cite{CAN}  & \emph{ResNet12}  & 63.08 $\pm$ 0.15 & 80.26 $\pm$ 0.12\\
MetaOptNet~\cite{metaoptnet}  & \emph{ResNet12}  & 65.99 $\pm$ 0.72  & 81.56 $\pm$ 0.53 \\
MatchNet~\cite{matchingnet}  & \emph{ResNet12}  & 68.50 $\pm$ 0.92 & 80.60 $\pm$ 0.71 \\
ProtoNet~\cite{protonet}  & \emph{ResNet12}  & 68.23 $\pm$ 0.23  & 84.03 $\pm$ 0.16 \\ %
CAN~\cite{CAN}  & \emph{ResNet12}  & 69.89 $\pm$ 0.51 & 84.23 $\pm$ 0.37\\
RFS-simple~\cite{rfs} & \emph{ResNet12}  & 71.61 $\pm$ 0.49 & 85.29 $\pm$ 0.24\\
\hline
 \textbf{MLCN(ours)}  &  \emph{ResNet12}  &  \textbf{71.62 $\pm$ 0.49} & \textbf{ 85.58} $\pm$ \textbf{0.35} \\ \hline
\end{tabular}
}

\end{table}

\begin{table}[t!]
\caption{Results on the CUB-200-2011 dataset.}
\centering
\small

\begin{tabular}{lccc}
\hline
\textbf{Mthod} & \textbf{Backbone} & \textbf{5-way 1-shot} & \textbf{5-way 5-shot}\\  
\hline
RelationNet~\cite{RelationNet} & \emph{ResNet18} & 68.58 $\pm$ 0.94 & 84.05 $\pm$ 0.56 \\
CloserLook~\cite{closer}  & \emph{ResNet18}  & 47.12 $\pm$ 0.74 & 64.16 $\pm$ 0.71\\
Baseline++\cite{closer}  & \emph{ResNet18}  & 67.02 $\pm$ 0.90 & 83.58 $\pm$ 0.50\\
MixtFSL~\cite{afrasiyabi2021mixture}  & \emph{ResNet18}  & 73.94 $\pm$ 1.10 & 86.01 $\pm$ 0.50\\
\hline
MAML~\cite{finn2017model} & \emph{ResNet34}${}^{\dag}$ & 67.28 $\pm$ 1.08 & 83.47 $\pm$ 0.59 \\	%
S2M2~\cite{s2m2} & \emph{ResNet34}${}^{\dag}$  & 72.92 $\pm$ 0.83 & 86.55 $\pm$ 0.51 \\
\hline
CC~\cite{closer} & \emph{ResNet12}  & 67.30 $\pm$ 0.86  & 84.75 $\pm$ 0.60 \\ %
ProtoNet~\cite{protonet} & \emph{ResNet12} & 66.09 $\pm$ 0.92 & 82.50 $\pm$ 0.58 \\
MatchNet~\cite{matchingnet} & \emph{ResNet12} & 71.87 $\pm$ 0.85 & 85.08 $\pm$ 0.57 \\
FEAT~\cite{feat}  & \emph{ResNet12}  & 73.27 $\pm$ 0.22 & 85.77 $\pm$ 0.14\\
DeepEMD~\cite{deepemd}  & \emph{ResNet12} &  75.65 $\pm$ 0.83 &  88.69 $\pm$ 0.50 \\
\hline
\textbf{MLCN(ours)}  &  \emph{ResNet12}  &  \textbf{77.96 $\pm$ 0.44} &  \textbf{91.20 $\pm$ 0.24} \\ 
\hline
\end{tabular}

\end{table}

\begin{table}[t!]
\caption{Results on the CIFAR-FS dataset.}
\centering
\tabcolsep 4pt
\small

\scalebox{0.98}{
\begin{tabular}{lccc}
\hline
\textbf{Mthod} & \textbf{Backbone} & \textbf{5-way 1-shot} & \textbf{5-way 5-shot}\\
\hline
S2M2~\cite{s2m2} & \emph{ResNet34}${}^{\dag}$  & 62.77 $\pm$ 0.23 & 75.75 $\pm$ 0.13 \\
MAML\cite{finn2017model} & \emph{ConvNet}  &58.90 $\pm$1.90  & 71.50 $\pm$ 1.00\\
\hline
DeepEMD\cite{deepemd} & \emph{ResNet12}  &46.47 $\pm$0.70  & 63.22 $\pm$ 0.71\\
R2D2~\cite{bertinetto2018meta}  & \emph{ResNet12}  & 65.30 $\pm$ 0.02 & 78.30 $\pm$ 0.02\\
RelationNet\cite{RelationNet} & \emph{ResNet12}  &55.50 $\pm$1.00  & 69.30 $\pm$ 0.80\\
CC~\cite{closer} & \emph{ResNet12}  & 60.39 $\pm$ 0.28 & 72.85 $\pm$ 0.65 \\ %
RFS-simple~\cite{rfs}  & \emph{ResNet12} & 71.50 $\pm$ 0.80  & 86.00 $\pm$ 0.50 \\
ProtoNet~\cite{protonet} & \emph{ResNet12}  & 72.20 $\pm$ 0.70  & 83.50 $\pm$ 0.50 \\ %
MetaOptNet~\cite{metaoptnet} & \emph{ResNet12}  & 72.60 $\pm$ 0.70  & 84.30 $\pm$ 0.50\\
\hline
 \textbf{MLCN(ours)}  &  \emph{ResNet12}  &  \textbf{74.36 $\pm$ 0.46} &  \textbf{87.24 $\pm$ 0.31} \\ 
\hline
\end{tabular}
}

\end{table}

\subsection{Datasets and Implementation Details}
\textbf{Datasets.} For experiment evaluation, we use four standard benchmarks for few-shot image classification: \textit{mini}ImageNet, \textit{tiered}ImageNet, CUB-200-2011 and CIFAR-FS. Following previous works\cite{CAN,kang2021renet}, we adopt the ResNet12 as our backbone, which consists of 4 residual blocks. The backbone network takes an image with spatial size of $84\times84$ as an input and provides a base representation $F\in R^{5\times5\times640}$ followed by shifting its channel activation by the  channel mean of an episode. For the N-way K-shot evaluation, we test 15 query samples for each class in an episode and report average classification accuracy with 95 \% confidence intervals of randomly sampled 2000 test episodes. The model is trained for 100 epochs.
For optimization, we use stochastic gradient descent (SGD) with a momentum of $0.9$, a weight decay of $5\times 10^{-4}$, a learning rate of $5\times 10^{-2}$.
We set the temperature scaling factor $\tau_{1}$ , $\tau_{2}$ and $\tau_{3}$ are set to 0.5 for all four datasets.
The ratio in the loss is set as $\alpha$:$\beta$:$\gamma$ = 4:2:1.

\subsection{Performance Comparison}
We compare the performance of MLCN with several state-of-the-art models including meta-learning and metric learning methods.
For fair comparisons, we compare with other methods in the same backbone or smaller backbone in both 5-way-1shot and 5-way-5shot settings on {\textit{mini}ImageNet}, {\textit{tired}ImageNet}, CUB-200-2011 and CIFAR-FS datasets. 
As is shown in Tables \textcolor{red}{2-5}, our method is superior to existing methods and achieves the best performance than both on meta-based methods and metric-based methods.

\subsection{Ablation Study and Visualization}
\textbf{Ablation Study.} In this subsection, we study the effectiveness of different components in our method on three datasets. 
All the results are summarized in Table \textcolor{red}{1}.
The results in Table 1 show a signification improvement in the performance of our proposed method compared to the baseline, which only uses the $\mathcal{L}_{CE}$.
Specifically, using three modules on the $\mathcal{L}_{CE}$ and $\mathcal{L}_{SC}$ improves the accuracy by an average of 7.2 \% (1-shot) and 8.6 \% (5-shot)  on the three datasets.
This allows the representation to generalize better than that only needed for the classification on $\mathcal{L}_{CE}$.
Then, we then use the $\mathcal{L}_{CC}$ and $\mathcal{L}_{PC}$ to capture local information of foreground images. The result jointly based on $\mathcal{L}_{CC}$ and $\mathcal{L}_{PC}$ are further enhanced by an average of 1.4 \% (1-shot) and 1.3 \% (5-shot) on the three datasets.
The results indicate that our method increases the transferability of embedding on novel classes.

\begin{figure}[t]
\centering
\includegraphics[width=0.9\linewidth]{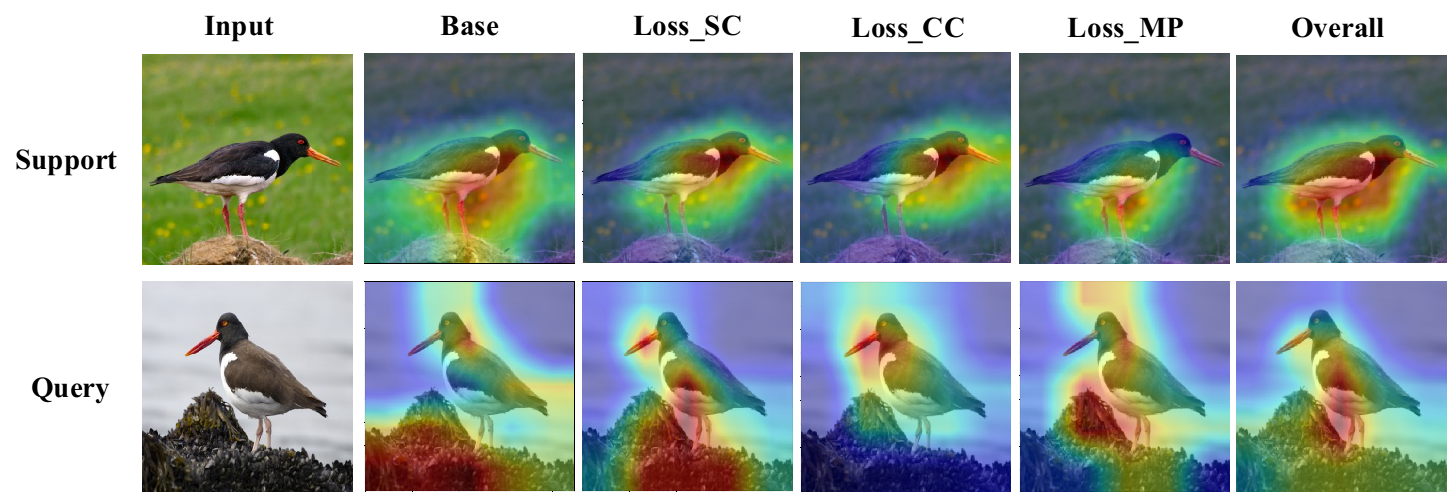}
\caption{GradCAM~\cite{selvaraju2017grad} visualization of the self-correlation module loss $\mathcal{L}_{SC}$, the cross-correlation module loss $\mathcal{L}_{CC}$ and the pattern-correlation loss $\mathcal{L}_{PC}$. The overall is the combined loss used in our MLCN.}
\label{fig:gradcam}
\end{figure}
\textbf{Visualization.} 
We give a visualization to validate the transferability of embedding produced by our framework on novel classes. Fig.\textcolor{red}{~\ref{fig:gradcam}} visualizes the gradient-weighted class activation mapping (Grad-CAM)~\cite{selvaraju2017grad} from four loss functions under a ResNet12 feature extractor. It is observed that using $\mathcal{L}_{SC}$ and $\mathcal{L}_{CC}$ pays more attention to the relevant salient object of the local information between query and support images. It further shows that using $\mathcal{L}_{PC}$ captures the relevant pattern of query and support images. Therefore, the proposed multi-level correlation network helps the metric-learning methods to use correct visual features.

\section{Conclusion}
In this paper, we propose multi-level correlation network (MLCN) for few-shot image classification (FSIC), which levrages the self-correlation module, cross-correlation module and pattern-correlation module.
By combining the three modules, our method effectively captures local information for FSIC. Extensive experiments demonstrate the effectiveness of our method on widely used FSIC benchmarks.

\vspace{12pt}

\clearpage

\title{Multi-Level Correlation Network for Few-Shot Image Classification}  

\maketitle
\thispagestyle{empty}
\appendix

\noindent We thank all reviewers for their constructive comments! We are excited to see that reviewers agree that our overall idea is interesting with good writing (\textcolor{YellowOrange}{\textbf{R\#1}}), sufficient experiments (\textcolor{red}{\textbf{R\#3}}). We address the main comments as below:

\noindent (\textcolor{YellowOrange}{\textbf{R\#1}}) \textcolor{blue}{Do hyper-parameters have a significant impact on model performance?}
We experimentally found that different hyper-parameters have no significant impact on model performance. More results are shown in Tables ~\ref{Tab:re2} and~\ref{Tab:re3}.  

\noindent (\textcolor{YellowOrange}{\textbf{R\#1}}) \textcolor{blue}{Are parameters provided in the paper always appropriate?} Yes. As shown in Table ~\ref{Tab:re2} and Table ~\ref{Tab:re3} , the hyper-parameters values used in our paper achieve the best performance. Moreover, the performance difference between different hyper-parameters choices is significantly small.


\noindent (\textcolor{red}{\textbf{R\#3}}) \textcolor{blue}{Compared with PMM method \& our innovative contributions.} 
In this paper, we propose our MLCN motivated by PMM with the following differences: \ding{182} 
PMM distinguishes foreground and background features by using a ground-truth mask. This cannot be applied to real-world data since it is difficult to label ground-truth mask. Instead, we use the output feature map based the self-correlation and cross-correlation module as the prior knowledge without the ground-truth mask.
\ding{183} PMM uses the Expectation-Maximization (EM) algorithm to divide positive and negative prototypes used for segmentation tasks. 
To save the computing time and memory brought by the EM algorithm, MLCN uses a first-order approximate bi-level algorithm. 

\noindent (\textcolor{red}{\textbf{R\#3}}) \textcolor{blue}{Ablation study.} 
The abilation study shows that the cross-correlation and pattern-correlation module improve the accuracy by 1.3 \% and 1.8 \%(5-shot) on the CUB-200-2011 dataset. 
And combining three modules can significantly improve the accuracy by 1.23 \% (1-shot) and 0.83 \% (5-shot) on \textit{mini}ImageNet and by 3.17 \% (1-shot) and 0.85 \% (5-shot) on CIFAR-FS.
It shows the potential ability of our method in few-shot image classification datasets.

\noindent (\textcolor{red}{\textbf{R\#3}}) \textcolor{blue}{Our motivation.}
In this paper, our main motivation is that image background is detrimental in transferring knowledge from training to testing data, \textit{i.e.}, reducing the generalization ability of the model. These results in Figure 1 also demonstrate our motivation. 
Therefore, we propose a multi-level correlation network to learn local information (including the foreground of the image).

\noindent (\textcolor{red}{\textbf{R\#3}}) \textcolor{blue}{The explanation of Figure 1.}
FG (or BG) means that we crop each image manually according to the largest rectangular bounding box that contains the foreground (or background) information.
In our experiment settings, for ``BG-FG", the former means training the model using only background information, and the latter means testing the model using only foreground information. 
Figure 1 shows that model trained with only foreground perform (\textit{i.e.}, ``FG-FG" or ``FG-BG") much better than those trained with background (\textit{i.e.}, ``BG-FG'' or ``BG-BG''). 
Background information at training serves as a shortcut for models to learn and cannot generalize from training data to testing data.

\noindent (\textcolor{red}{\textbf{R\#3}}) \textcolor{blue}{About references.} 
Thanks for pointing out the references\noindent{1}, \noindent{2}, we will add them in the revised version. 
Self-attention relation network use the attention module to highlight the target object.
Different from [1], we use the self-correlation module to learn local feature with less parameters.
SaberNet [2] use the vision transformer to extract local feature by splitting each image into patches as the input sequence and use the self-attention mechanism based patches. 
Different from [2], we use the ResNet12 architecture to extract feature without patches to reduce the computing complexity and avoid overfitting for few-shot classification. 

\noindent (\textcolor{red}{\textbf{R\#3}}) \textcolor{blue}{About symbolic explanation.}
We are sorry for not making this clear and will edit them in the revised version. 
On an N-way K-shot classification setting, this self-correlation module generates a set of NK different views of a query, $\left \{\mathbf{z_{q}}^{(n)} \right \}_{n=1}^{NK}$, and a set of support embeddings attended in the context of the query, $\left \{\mathbf{z_{s}}^{(n)} \right \}_{n=1}^{NK}$. 
Before computing the loss, we average the K query embedding vectors and each of which is attended in the context of $k^{th}$ support from $n^{th}$ class to compute $\left \{\mathbf{z_{q}}^{(n)} \right \}_{n=1}^{N}$. Similarly, we average the K support embeddings for each class to obtain a set of prototype embeddings $\left \{\mathbf{z_{s}}^{(n)} \right \}_{n=1}^{N}$.

\noindent (\textcolor{red}{\textbf{R\#3}}) \textcolor{blue}{About symbol representation and english grammar.}
Thank you for your valuable comments, which help us to improve the quality of this paper. We apologize for the reading distress caused by unclear symbol representation. 
We will unify the symbol representation and adjust the English grammar in the newly edited manuscript.

\begin{table}[tbp]
    \caption{Results for different hyper-parameters on \textit{mini}Imagenet, CUB-200-2011, and CIFAR-FS datasets under 5-way 5-shot.}
    \begin{center}
        \resizebox{1.0\linewidth}{!}{
            \begin{tabular}{l|cccc}
                \toprule
                $\alpha : \beta : \gamma$ & \textit{mini}ImageNet & CUB-200-2011 & CIFAR-FS & \textit{tiered}ImageNet \\
                \midrule
                 4 : 4 : 1 & 81.73 ± 0.39 &90.44 ± 0.25 &  86.97 ± 0.32 & 85.12 ± 0.35 \\
                 4 : 2 : 1 & 81.94 ± 0.31 &91.20 ± 0.24 &  87.24 ± 0.31 & 85.58 ± 0.35\\
                 2 : 2 : 1 & 81.31 ± 0.31 &89.72 ± 0.26 &  86.67 ± 0.33 & 85.27 ± 0.35 \\
                 2 : 1 : 1 & 81.38 ± 0.32 &89.82 ± 0.26 &  86.36 ± 0.32 & 85.29 ± 0.34\\
                \bottomrule
            \end{tabular}}
    \end{center}
\vspace{-19pt}
\label{Tab:re2}
\end{table}

\begin{table}[tbp]
    \caption{Results for different hyper-parameters on CUB-200-2011 dataset under 5-way 1-shot and  5-way 5-shot.}
    \begin{center}
        \resizebox{0.5\linewidth}{!}{
            \begin{tabular}{l|cccc}
                \toprule
                $\tau$ & 5-way 1-shot & 5-way 5-shot \\
                \midrule
                 0.1 & 77.31 ± 0.43 &  90.72 ± 0.25  \\
                 0.3 & 77.73 ± 0.44 &  91.01 ± 0.24\\
                 0.5 & 77.96 ± 0.44 &  91.20 ± 0.24 \\
                 0.7 & 77.63 ± 0.39 &  90.92 ± 0.25\\
                 0.9 & 77.62 ± 0.43 &  90.85 ± 0.25\\
                \bottomrule
            \end{tabular}}
    \end{center}
\vspace{-19pt}
\label{Tab:re3}
\end{table}

\noindent 1.Hui et al. Self-attention relation network for few-shot learning. ICME Workshops (ICMEW), 2019.

\noindent 2.Li et al. SaberNet: Self-attention based effective relation network for few-shot learning. Pattern Recongnition, 2023.

\end{document}